%% file: neurips_2026.tex
\documentclass{article}

\PassOptionsToPackage{numbers, compress}{natbib}

\usepackage[preprint]{neurips_2026}

\usepackage[utf8]{inputenc}
\usepackage[T1]{fontenc}
\usepackage{hyperref}
\usepackage{url}
\usepackage{booktabs}
\usepackage{amsfonts}
\usepackage{nicefrac}
\usepackage{microtype}
\usepackage{graphicx}
\usepackage{natbib}
\usepackage{multirow}
\usepackage{adjustbox}
\usepackage{xcolor}
\usepackage{colortbl}
\usepackage{amsmath,amssymb}
\usepackage{amsthm}

\usepackage{algorithm}
\usepackage{algorithmic}

\usepackage{subcaption}
\usepackage{wrapfig}
\usepackage{enumitem}
\usepackage{comment}

\usepackage[capitalize,noabbrev]{cleveref}

\definecolor{posblue}{HTML}{1A6FB5}
\definecolor{negred}{HTML}{C0392B}
\definecolor{lightgray}{gray}{0.92}

\newtheorem{theorem}{Theorem}[section]
\newtheorem{proposition}[theorem]{Proposition}

\theoremstyle{definition}

\newtheorem{definition}[theorem]{Definition}

\theoremstyle{remark}

\graphicspath{{figures/}}

\newcommand{\E}{\mathbb{E}}
\newcommand{\R}{\mathbb{R}}
\newcommand{\Prob}{\mathbb{P}}

\newcommand{\cM}{\mathcal{M}}

\newcommand{\cA}{\mathcal{A}}
\newcommand{\cO}{\mathcal{O}}
\newcommand{\cH}{\mathcal{H}}

\newcommand{\vs}{\mathbf{s}}

\newcommand{\va}{\mathbf{a}}
\newcommand{\vo}{\mathbf{o}}
\newcommand{\vd}{\mathbf{d}}

\newcommand{\doop}{\mathrm{do}}
\newcommand{\SoI}{\mathrm{SoI}}
\newcommand{\IBD}{\textsc{IBD}}

\title{Discovering What You Can Control: \\
Interventional Boundary Discovery \\ for Reinforcement Learning}

\author{%
  Jiaxin Liu \\
  University of Illinois Urbana-Champaign \\
  \texttt{jiaxin26@illinois.edu} \\
  \And
  Anzhe Cheng \\
  University of Southern California \\
  \texttt{anzheche@usc.edu} \\
  \And
  Paul Bogdan \\
  University of Southern California \\
  \texttt{pbogdan@usc.edu} \\
}

\begin{document}

\maketitle


\begin{abstract}
When an RL agent's observations contain distractors driven by the same confounders as its true state, observational data alone cannot identify which dimensions the agent controls.
In our benchmarks, even state-conditioned observational selectors can collapse when distractors mimic controllable state variables.
We propose Interventional Boundary Discovery (\IBD{}), which treats the agent's own action channel as a source of randomized interventions: randomizing actions implements an interventional contrast, and per-dimension two-sample tests with FDR correction produce a binary mask over observation dimensions.
Across 12 continuous-control settings with up to 100 distractors, \IBD{} matches oracle return in 11 of 12 settings, while observational baselines including mutual information, state-conditioned forward models, and gradient-based sensitivity often underperform simply passing the full observation to SAC. Code is available at \url{https://github.com/jiaxin26/IBD-RL}
\end{abstract}


\section{Introduction}
\label{sec:intro}

Reinforcement Learning (RL) has achieved strong results in continuous control and sequential decision making~\citep{kang2025forget, zhang2024wococo}, with modern agents solving increasingly challenging tasks from robotics to simulated benchmarks~\citep{haarnoja2018sac,fujimoto2018td3,tassa2018dmcontrol}. A common assumption behind this progress is that the agent receives Markov or near-Markov observations that are sufficiently informative about the underlying state~\citep{jeen2025zero,nguyen2024leveraging,huang2025pigdreamer}. In practice, observations often mix truly controllable state variables with many irrelevant dimensions (nuisance signals, background dynamics, sensor artifacts, and exogenous processes), a problem widely recognized in both visual RL~\citep{stone2021distracting,hansen2021svea,yarats2022drqv2,laskin2020curl} and state-space RL~\citep{efroni2022provably,wang2022denoised}. Such distractors are especially harmful when they are not merely random noise, but statistically resemble the true state or correlate with the agent's behavior. In this case, a policy trained on the full observation can spend capacity modeling non-causal variation, while a passive feature selector can retain dimensions that look predictive but are outside the agent's control.

This failure is not merely a consequence of high-dimensional inputs. The more difficult case arises when irrelevant dimensions are statistically entangled with the agent's behavior. For example, an exogenous coordinate may change over the same time scale as the task state, correlate with episode progress, or share latent causes with the agent's actions~\citep{wang2022denoised}. Such a dimension can therefore appear useful under passive observation, even though intervening on the agent's action would not change its distribution. Conversely, a truly action-reachable dimension may have a weaker observational association if its effect is delayed, stochastic, or only visible under particular action sequences. Thus, passive feature selection can confuse correlation with controllability: it may retain dimensions that co-vary with actions while discarding dimensions that are actually downstream of actions. This creates an identifiability gap for observation selection in RL, where the goal is not only to predict the next observation but to identify which parts of the observation lie within the agent's causal reach.

Existing methods do not directly resolve this distinction (\Cref{sec:related}). Visual-RL methods such as SVEA, DrQ-v2, and CURL~\citep{hansen2021svea,yarats2022drqv2,laskin2020curl} suppress pixel-level nuisance variation but do not test whether a retained feature is causally influenced by actions, and exogenous-block and Denoised MDPs~\citep{efroni2022provably,wang2022denoised,levine2025exbmdp,levine2025craft} still compute their selection signal from observational trajectories, where a distractor that co-varies with the true state can look like a controllable dimension. Empowerment and causal representation learning~\citep{scholkopf2021causal,yao2025unifying,varici2025scorecrl} either aggregate action influence or target richer latent structure, and neither answers which observed dimensions actions can reach.

We propose \textbf{Interventional Boundary Discovery (IBD)}, a model-free procedure for identifying the agent's \emph{Causal Sphere of Influence} (SoI): the set of observed dimensions whose future distribution changes under an intervention on the agent's actions. IBD uses the agent's own action channel as an intervention mechanism. During a short probing phase, it randomizes actions to approximate an action intervention, compares baseline and intervention trajectory statistics with per-dimension two-sample tests, and applies false-discovery-rate (FDR) control~\citep{benjamini1995fdr}. The result is a binary mask over observation dimensions. This mask can preprocess the input to downstream RL algorithms such as SAC~\citep{haarnoja2018sac} and TD3~\citep{fujimoto2018td3}, without requiring a learned world model, latent graph recovery, or changes to the policy optimizer.
IBD also provides a diagnostic view of when feature selection matters. Comparing downstream performance under the Full State, IBD mask, and Oracle mask separates different failure modes. If IBD closes the Full State--Oracle gap, the main bottleneck is distractor-induced; if the gap is small, feature selection is not the limiting factor; if all three fail, the bottleneck lies beyond observation masking. Across 12 continuous-control settings, IBD recovers near-oracle masks and improves downstream learning when observational baselines are misled by spuriously correlated distractors, and it stays accurate when a genuine latent confounder is added (\Cref{sec:confounding}).

The primary contributions of this work are summarized as follows:
\begin{itemize}
    \item We formalize the \emph{Causal Sphere of Influence} in RL and show that observational feature selection has a fundamental identifiability gap: confounded distractors can rank above truly action-reachable dimensions.
    \item We introduce \textbf{Interventional Boundary Discovery}, a model-free procedure using randomized action interventions, per-dimension hypothesis testing, and FDR control to recover a binary mask over controllable dimensions.
    \item Across 12 continuous-control settings, IBD delivers near-oracle downstream RL performance where observational baselines often fail, while also serving as a diagnostic for separating distractor-induced failures from other RL bottlenecks.
\end{itemize}

\section{Related Work}
\label{sec:related}

\textbf{Distractor-Robust Reinforcement Learning.} Distractors degrade RL agents through both representation learning and optimization difficulty~\citep{stone2021distracting}. In visual RL, CURL~\citep{laskin2020curl}, SVEA~\citep{hansen2021svea}, and DrQ-v2~\citep{yarats2022drqv2} reduce sensitivity to pixel-level nuisance variation through contrastive objectives and strong augmentation. In state-space RL, Denoised MDPs~\citep{wang2022denoised} and the Exogenous Block MDP framework~\citep{efroni2022provably}, with extensions to single-trajectory and action-free settings~\citep{levine2025exbmdp,levine2025craft}, separate endogenous from exogenous factors via inverse dynamics or latent decomposition. These methods improve robustness, but their selection signal is computed from observational trajectories: a distractor whose dynamics co-vary with action-reachable state can still appear useful for inverse dynamics or forward prediction even when no intervention on the action affects it. \IBD{} replaces this passive signal with an active interventional contrast that severs confounding paths at the action node.

\textbf{Causal Representation Learning.} Causal representation learning aims to recover latent causal variables from temporal observations~\citep{scholkopf2021causal}. CITRIS~\citep{lippe2022citris} and iCITRIS~\citep{lippe2023icrl} identify latent variables under known intervention targets, with extensions to invariance principles~\citep{yao2025unifying} and broader intervention classes~\citep{varici2025scorecrl,varici2024multinode}. These methods recover a richer object than feature selection requires (a latent graph rather than a mask over observed dimensions), typically under assumptions on intervention labels, temporal variation, or latent transition structure. \IBD{} targets the narrower question of dimension-level action reachability in observed space, requires no latent recovery, and can serve as a pre-filter before more detailed graph-level analysis.

\textbf{Control, Empowerment, and Feature Selection.} Empowerment~\citep{klyubin2005empowerment,mohamed2015variational} measures channel capacity between actions and future states, but as an aggregate scalar rather than a dimension-level criterion: high $I(A;S'\!\mid\!S)$ does not imply that a specific observed dimension is causally affected by actions. Classical controllability analysis~\citep{kalman1960controllability} provides a precise model-based notion of reachability for linear systems with known dynamics; recent work on interpretable controllability features~\citep{kooi2023controllable} routes through learned latent representations rather than direct observed-space testing. Classical feature-selection criteria (mutual information, variance, forward-model sensitivity) form natural baselines but conflate statistical and causal relevance: a distractor highly correlated with actions through shared latent causes or behavioral progress can rank highly without being causally influenced. State abstraction~\citep{li2006towards} groups states by optimal-policy equivalence, a different objective. \IBD{} addresses dimension-level action reachability directly through interventional contrasts in the observed space, requiring neither a known model nor a learned latent representation.

\section{Method}
\label{sec:method}


\begin{figure}[t]
\centering
\includegraphics[width=\linewidth]{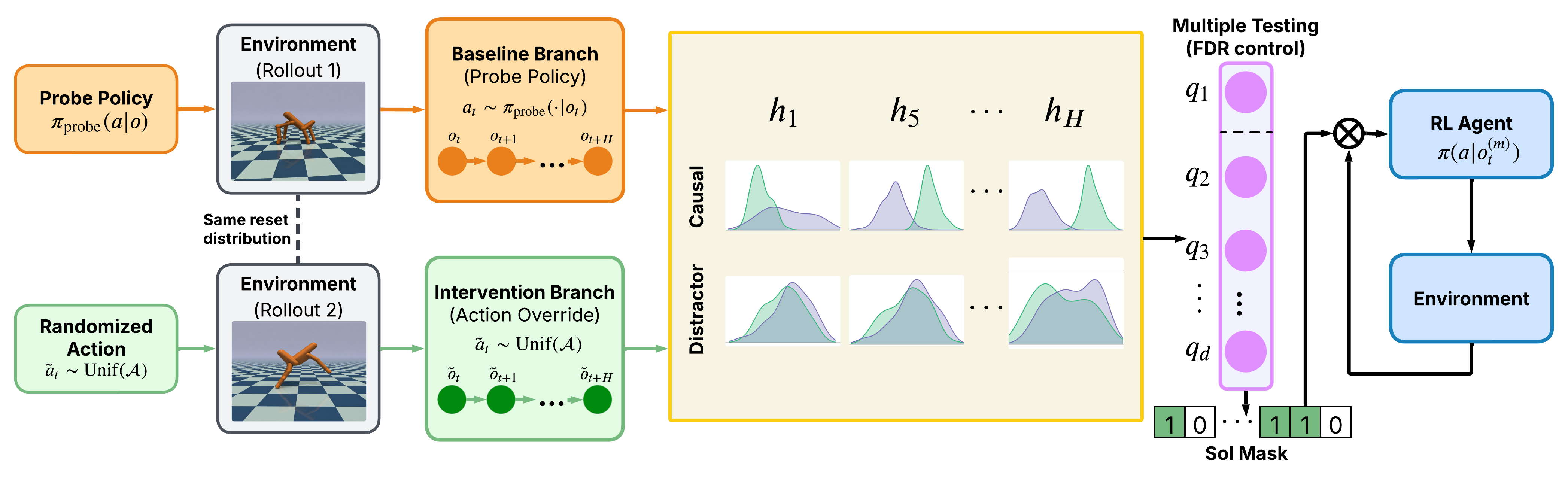}
\caption{\textbf{Overview of Interventional Boundary Discovery (IBD).}
From the same reset distribution, IBD collects two independent rollouts: a baseline branch under a probe policy $\pi_{\mathrm{probe}}$, and an intervention branch in which the action mechanism is replaced by a confounder-independent randomized policy, $\tilde a_t \sim \mathrm{Unif}(\mathcal{A})$.
For each observation dimension and horizon, IBD compares trajectory statistics across the two branches using a two-sample test. Benjamini--Hochberg FDR correction then yields a binary SoI mask, which filters observations before downstream RL training.}
\label{fig:pipeline}
\vspace{-5mm}
\end{figure}

\subsection{The Causal Sphere of Influence}
\label{sec:formulation}

We consider a Markov Decision Process $\cM = (\cO, \cA, T, R, \gamma)$ where the observation space $\cO \subseteq \R^d$ decomposes (unknown to the agent) as $\vo_t = [\vs_t^{\mathrm{c}}, \vs_t^{\mathrm{d}}]$: a causal component $\vs^{\mathrm{c}} \in \R^{d_c}$ that is causally downstream of the agent's actions $\va_t \in \cA \subseteq \R^{d_a}$, and a distractor component $\vs^{\mathrm{d}} \in \R^{d_d}$ that evolves autonomously.
The total observation dimension is $d = d_c + d_d$, and the agent does not know which dimensions are causal.

\begin{definition}[Causal Sphere of Influence]
\label{def:soi}
Observation dimension $i$ belongs to the agent's \emph{Sphere of Influence} ($\SoI$) if and only if there exists an action dimension $j$ and a horizon $h \geq 1$ such that the intervention $\doop(a^{(j)}_t = u)$ changes the marginal distribution of $o^{(i)}_{t+h}$:
\begin{equation}
\label{eq:soi}
\Prob\!\left(o^{(i)}_{t+h} \,\middle|\, \doop(a^{(j)}_t = u)\right) \;\neq\; \Prob\!\left(o^{(i)}_{t+h}\right)
\end{equation}
for some intervention value $u$, where $\doop(\cdot)$ denotes Pearl's do-operator~\citep{pearl2009causality}.
\end{definition}

The definition says that dimension $i$ is inside the SoI if and only if there exists a directed causal path from some action to dimension $i$ in the environment's causal graph.
Note that \eqref{eq:soi} involves the do-operator, not conditional probability.
A dimension correlated with actions due to a common confounder satisfies $\Prob(o^{(i)} | a) \neq \Prob(o^{(i)})$ but does not satisfy $\Prob(o^{(i)} | \doop(a = u)) \neq \Prob(o^{(i)})$, because the do-operator severs the confounding path.

The goal of \IBD{} is to recover $\SoI$ from interaction data, producing a binary mask $\mathbf{m} \in \{0, 1\}^d$ with $m_i = \mathbf{1}[i \in \widehat{\SoI}]$.
The downstream RL algorithm then receives only the masked observation $\vo_t \odot \mathbf{m}$.

\paragraph{Assumptions.}
\IBD{} relies on four assumptions:
(i) decomposability: dimensions are either fully causal or fully exogenous (relaxed in \Cref{sec:robustness} to partially controllable dimensions);
(ii) action overrideability during probing, which is natural in simulation and requires a safe-exploration protocol on hardware;
(iii) stationary causal structure during probing;
(iv) faithfulness and statistic sensitivity: a directed action-to-dimension path produces a non-trivial distributional shift that is detectable by the chosen summary $\Delta^h_i$ for some $h \in H$ (Proposition~\ref{prop:detect}).
The default $\Delta^h_i$ is insensitive to dimensions that are action-reachable only through a shift in mean level; taking the union of $\Delta^h_i$ tests and per-trajectory mean tests recovers these (Appendix~\ref{app:summary_stat}).

\subsection{Algorithm}
\label{sec:algorithm}

\IBD{} estimates the SoI by an interventional contrast: trajectories collected under a probe policy are compared, dimension by dimension, against trajectories in which the action channel is randomized. Dimensions outside the SoI remain distributionally invariant under this contrast up to sampling noise, while descendants of actions exhibit a measurable distributional shift at some horizon.

From the same reset distribution we collect $N$ baseline trajectories $\{\tau^{\mathrm{b}}_k\}_{k=1}^{N}$ under probe policy $\pi_{\mathrm{probe}}$, and $N$ intervention trajectories $\{\tau^{\mathrm{int}}_k\}_{k=1}^{N}$ in which each action is replaced by an i.i.d.\ draw from $\mathrm{Unif}(\cA)$. This implements a stochastic intervention by replacing the behavior-policy mechanism with the confounder-independent distribution $q(\va)=\mathrm{Unif}(\cA)$; Uniform is maximum-entropy on $\cA$ and maximally excites action-reachable dimensions, but any confounder-independent distribution is valid. Our default $\pi_{\mathrm{probe}}$ is a structured random policy with sinusoidal actions and weak state feedback, requiring no RL training (\Cref{sec:practical}).

For each dimension $i$ and horizon $h \in \cH$ (we use $\cH = \{1, 5, 10\}$), each trajectory is summarized by the mean absolute $h$-step difference,
\begin{equation}
\label{eq:summary}
\Delta^h_i(\tau) \;=\; \frac{1}{K_h} \sum_{r=0}^{K_h - 1} \left| o^{(i)}_{(r+1)h} - o^{(i)}_{rh} \right|, \qquad K_h = \lfloor T / h \rfloor,
\end{equation}
yielding $N$ scalars per branch that are i.i.d.\ across trajectories. Under this randomized-action intervention, an exogenous dimension has an action-invariant transition kernel and $\Delta^h_i$ has the same distribution under both branches; an action-reachable dimension exhibits a shifted distribution. Multiple horizons are tested because a causal chain of length $k$ may produce no detectable effect until $h \geq k$. Since the final object is a dimension-level mask rather than a list of $(i, h)$ discoveries, we aggregate the $|\cH|$ horizon-level $p$-values within each dimension via the Bonferroni minimum $\tilde p_i = \min\{1,\ |\cH| \min_{h \in \cH} p_{i,h}\}$, then apply Benjamini--Hochberg correction~\citep{benjamini1995fdr} across the $d$ dimension-level $p$-values at level $\alpha = 0.05$. The two samples are compared with a Welch $t$-test. The output is a binary mask $\mathbf{m} \in \{0, 1\}^d$ with $m_i = \mathbf{1}[q_i < \alpha]$, computed once at a cost of $2NT$ environment steps and reused throughout downstream training. \Cref{alg:ibd} summarizes the procedure.

\begin{algorithm}[H]
\caption{Interventional Boundary Discovery (\IBD{})}
\label{alg:ibd}
\begin{algorithmic}[1]
\REQUIRE Environment $\cM$, probe policy $\pi_{\mathrm{probe}}$ (default: structured random); trajectory count $N$ and length $T$ (default $N{=}80$, $T{=}200$); horizons $\cH$ (default $\{1, 5, 10\}$); significance $\alpha$ (default $0.05$)
\STATE Collect baseline $\{\tau^{\mathrm{b}}_k\}_{k=1}^{N}$ using $\pi_{\mathrm{probe}}$ and intervention $\{\tau^{\mathrm{int}}_k\}_{k=1}^{N}$ using $\doop(\va \sim \mathrm{Unif}(\cA))$
\FOR{each dimension $i \in [d]$}
    \FOR{each horizon $h \in \cH$}
        \STATE Compute $\{\Delta^h_i(\tau^{\mathrm{b}}_k)\}_{k=1}^{N}$ and $\{\Delta^h_i(\tau^{\mathrm{int}}_k)\}_{k=1}^{N}$ via Eq.~\ref{eq:summary}
        \STATE Obtain $p_{i,h}$ from a Welch $t$-test on the two samples
    \ENDFOR
    \STATE $\tilde p_i \leftarrow \min\{1,\ |\cH| \min_{h \in \cH} p_{i,h}\}$ \hfill (Bonferroni over horizons)
\ENDFOR
\STATE Apply Benjamini--Hochberg to $\{\tilde p_i\}_{i=1}^{d}$ at level $\alpha$, yielding $\{q_i\}_{i=1}^{d}$
\RETURN Binary mask $\mathbf{m}$ with $m_i = \mathbf{1}[q_i < \alpha]$
\end{algorithmic}
\end{algorithm}
\vspace{-6mm}

\subsection{Failure of Observational Selection}
\label{sec:why_obs_fail}
\vspace{-3mm}
Before presenting the theoretical guarantees of \IBD{}, we give a construction illustrating a failure mode of observational criteria.
Natural dependence-based criteria, mutual information in particular, can fail to separate a confounded distractor from the true causal dimension it tracks, regardless of how well the dependence is estimated.

\begin{proposition}[Observational dependence can rank confounded distractors above causal dimensions]
\label{prop:obs_fail}
A distractor can have higher mutual information with behavior actions than a causal state dimension, even though actions have zero causal effect on it.
\end{proposition}

\emph{Construction.}
Consider the linear-Gaussian SCM
\[
C_t \sim \mathcal{N}(0,1),\qquad
a_{1,t}=C_t+\xi_t,\qquad
s_{1,t+1}=\beta a_{1,t}+\eta_t,\qquad
d_{k,t}=C_t+\varepsilon_t,
\]
where $\beta\neq 0$, with $\xi_t,\eta_t,\varepsilon_t$ independent noises. The distractor has no incoming edge from $\va_t$ or $s_t$; future distractors are generated autonomously by the same exogenous mechanism. Since both $a_{1,t}$ and $d_{k,t}$ share the latent cause $C_t$, $\mathrm{MI}(d_{k,t};\va_t)$ can exceed $\mathrm{MI}(s_{1,t+1};\va_t)$ by taking $\sigma_\xi,\sigma_\varepsilon$ small and the transition noise in $s_1$ larger. Thus an MI selector can rank $d_k$ above the causal dimension.

However, intervening on the action severs the edge $C_t\to a_{1,t}$ but leaves
the autonomous mechanism for $d_k$ unchanged. Hence, for every $h\ge 1$ and
intervention value $u$,
\[
P(d_{k,t+h}\mid \doop(\va_t=u))=P(d_{k,t+h}),
\]
whereas $P(s_{1,t+1}\mid \doop(a_{1,t}=u))$ changes with $u$ whenever
$\beta\neq 0$. Therefore observational dependence alone does not identify the
causal Sphere of Influence. \qed

This construction isolates the key failure mode: a dimension can share a latent cause with the behavior action and therefore appear highly action-informative in observational data, while remaining outside the causal downstream of any action intervention. Our benchmark (\Cref{sec:setup}) instantiates related patterns: mimicking dimensions whose marginal statistics match true proprioceptive channels, and reward-correlated dimensions that drift with episode progress. In both cases, the distractors can be statistically predictive under the behavior distribution without lying on a causal path from actions to future observations. \Cref{sec:results} shows that observational baselines, including a state-conditioned forward model, drop to near-random return on the harder regimes, while \IBD{} does not. \Cref{sec:confounding} instantiates the construction itself with a genuine latent confounder.
\vspace{-4mm}
\subsection{Identification Properties}
\label{sec:theory}
\vspace{-3mm}
The two-sample test in \IBD{} inherits standard properties from causal inference and multiple testing: the confounding path is severed at the action node, a causal path still produces a detectable shift under faithfulness, and FDR control bounds the rate of spurious inclusions.

\begin{proposition}[Invariance to confounders]
\label{prop:confound}
Because \IBD{} collects interventional data by replacing the action mechanism with a confounder-independent randomized policy, all edges $C \to \va$ from confounders $C$ to actions are severed under the intervention branch.
In the population limit, the distribution of the test statistic on non-SoI dimensions is therefore invariant to the presence or absence of confounders, and $\widehat{\SoI}$ converges to the same set whether confounders are active or removed (up to sampling noise in finite samples).
\end{proposition}

\begin{proposition}[Interventional detectability]
\label{prop:detect}
Assume (i) faithfulness: if a directed causal path from $a_j$ to $o_i$ of length $\leq h$ exists, then the conditional distribution of $o^{(i)}_{t+h}$ given $\doop(a_j = u)$ depends non-trivially on $u$; and (ii) statistic sensitivity: this distributional change is not annihilated by the chosen summary $\Delta^h_i$.
Then a directed causal path of length $\leq h$ from $a_j$ to $o_i$ is a sufficient condition for the distribution of $\Delta^h_i$ to differ between the baseline and intervention branches.
Conversely, in the absence of any such path, $\Delta^h_i$ has the same distribution under both branches.
\end{proposition}

\begin{proposition}[Type I and FDR control]
\label{prop:type1}
The Welch $t$-test on trajectory-level summaries $\{\Delta^h_i(\tau^b_k)\}, \{\Delta^h_i(\tau^{\mathrm{int}}_k)\}$ controls per-pair Type I error asymptotically under the central limit theorem. After Bonferroni aggregation over horizons within each dimension, $\tilde p_i = \min\{1, |H|\min_{h\in H} p_{i,h}\}$, the resulting dimension-level $p$-values are valid under the null. Applying Benjamini--Hochberg correction to $\{\tilde p_i\}_{i=1}^d$ controls the dimension-level false discovery rate at level $\alpha$ under the standard independence or PRDS dependence conditions.
\end{proposition}

The PRDS condition underlying Benjamini--Hochberg need not hold when mimicking distractors are statistically coupled to true dimensions; replacing BH with the Benjamini--Yekutieli procedure~\citep{benjamini2001dependency}, which is valid under arbitrary dependence, costs no F1 on our benchmark (Appendix~\ref{app:multiplicity}).

These results address the two failure modes of selection:
distractor leakage is controlled by FDR, while missed causal dimensions vanish asymptotically under faithfulness when the tested horizons cover the relevant causal delay and the intervention effect is detectable.

\vspace{-4mm}
\subsection{Practical Considerations}
\label{sec:practical}
\vspace{-3mm}
The probe policy need not be a trained RL agent; its role is to induce sufficient state coverage so that the intervention's causal signal is detectable. Our default is the structured random policy of \Cref{sec:algorithm}, which sufficed for all tasks in our experiments. A scout-budget sweep (Appendix~\ref{app:scout_ablation}) shows identical boundary accuracy on \texttt{cheetah\_run} and \texttt{reacher\_hard} between the zero-budget structured probe and an $80$K-step SAC scout, and higher F1 for the untrained probe on \texttt{walker\_walk} ($0.94$ vs.\ $0.90$), indicating that IBD does not require a learned exploratory policy as a prerequisite.
With our default settings ($N = 80$, $T = 200$), the probing phase requires approximately $32$K environment steps and is a one-time cost: the mask is computed once and reused for the entire downstream training run.

\input{table1_main}

\section{Experiments}
\label{sec:results}
\vspace{-4mm}
We evaluate whether randomized action interventions recover controllable observation dimensions when passive correlations are misleading. The benchmark appends distractors to continuous-control observations without changing the true state, reward, action space, or transition dynamics. This design lets us separate two questions: whether distractors degrade downstream RL, and whether the selected mask recovers the action-reachable boundary.
\vspace{-4mm}
\subsection{Benchmarks and Evaluation Protocol}
\label{sec:setup}
\vspace{-3mm}
We build a controlled state-space benchmark on six DeepMind Control Suite tasks using proprioceptive observations: \texttt{walker\_walk} (24 true dimensions), \texttt{cheetah\_run} (17), \texttt{hopper\_hop} (15), \texttt{finger\_spin} (9), \texttt{reacher\_hard} (6), and \texttt{cartpole\_swingup} (5). The original state dimensions define the ground-truth action-reachable set; appended distractors do not alter the transition, reward, or action dynamics, so the state-only observation is a valid oracle reference. We use three distractor regimes: autonomous OU/oscillator processes, mimicking processes whose scale, variance, and autocorrelation match true proprioceptive channels, and reward-correlated drift processes that track episode progress without being caused by actions. Easy adds 6 autonomous distractors; medium adds 50 and hard adds 100, with all three types present in medium and hard. 

After that, we compare Full State, Oracle, MI Select, Variance Select, Cond.\ MI, and \IBD{}. To favor observational selection, MI/Variance/Cond.\ MI receive the true number of causal dimensions $d_c$ as their selection budget; \IBD{} receives no budget hint and selects via its FDR-controlled significance threshold. Unless stated otherwise, all downstream agents use SAC with identical hyperparameters (lr $3\times 10^{-4}$, batch 256, replay buffer 300K, MLP [256,256], 300K environment steps). We report final episodic return as mean $\pm$ std over 5 seeds, with 10 evaluation episodes per seed, and report mask quality using precision, recall, and F1.






\vspace{-5mm}
\begin{table}[H]
\centering
\small
\caption{Gradient attribution baseline (episode return, mean $\pm$ std over 5 seeds, medium distractors). Grad.\ Attr.\ performs comparably to or worse than Cond.\ MI, confirming that the failure of observational methods under spurious correlation is not resolved by gradient-based sensitivity analysis.}
\label{tab:grad_attr}
\begin{tabular}{lccccc}
\toprule
Environment & Full State & Cond.\ MI & Grad.\ Attr. & \IBD{} (ours) & Oracle \\
\midrule
\texttt{cartpole\_swingup} & 725$\pm$48 & 161$\pm$60 & 56$\pm$32 & \textbf{834}$\pm$26 & 821$\pm$35 \\
\texttt{finger\_spin}      & 365$\pm$13 & 4$\pm$6    & 62$\pm$113 & \textbf{587}$\pm$178 & 775$\pm$181 \\
\texttt{hopper\_hop}       & 0$\pm$1    & 0$\pm$0    & 0$\pm$0   & \textbf{6}$\pm$12 & 6$\pm$8 \\
\texttt{reacher\_hard}     & 12$\pm$7   & 7$\pm$6    & 5$\pm$10  & \textbf{929}$\pm$76 & 925$\pm$75 \\
\bottomrule
\end{tabular}
\vspace{-4mm}
\end{table}

\vspace{-4mm}
\subsection{Main Control Results}
\vspace{-3mm}

Table~\ref{tab:main} gives the main downstream control results. \IBD{} reaches oracle-level performance in $11$ of $12$ evaluated settings, while passive feature selection often performs worse than training on the full noisy observation. This pattern is strongest in the medium and hard regimes, where distractors share state-like temporal statistics or progression-linked drift. These results support the central claim: the failure mode is not high dimensionality alone, but high-dimensional input whose passive correlations make non-causal dimensions look relevant. For example, for \texttt{reacher\_hard} medium, Full State reaches only $12\pm7$, and the three passive selectors remain near random: MI Select obtains $12\pm5$, Variance Select obtains $7\pm5$, and Cond.\ MI obtains $7\pm6$. In contrast, \IBD{} reaches $929\pm76$, matching Oracle at $925\pm75$. The gap is notable because Cond.\ MI uses a state-conditioned forward model to estimate whether actions add predictive information beyond the current state. Its collapse shows that conditioning can reduce marginal association but still cannot remove spurious dependence created by action-correlated distractor dynamics.

The same pattern appears beyond a single task. On \texttt{cheetah\_run} medium, Full State drops to $113\pm34$, while \IBD{} reaches $479\pm95$ and Oracle reaches $472\pm35$. On \texttt{finger\_spin} medium, all passive baselines collapse to near-zero return, while \IBD{} recovers $587\pm178$. These gains occur even though the passive selectors know the correct selection budget. Thus, the budget is not the main issue: the ranking signal is wrong when non-causal dimensions share passive statistics with causal state dimensions. \IBD{} avoids this failure by asking whether each dimension changes under randomized action intervention.

\begin{figure}[t]
\centering
\includegraphics[width=0.9\textwidth]{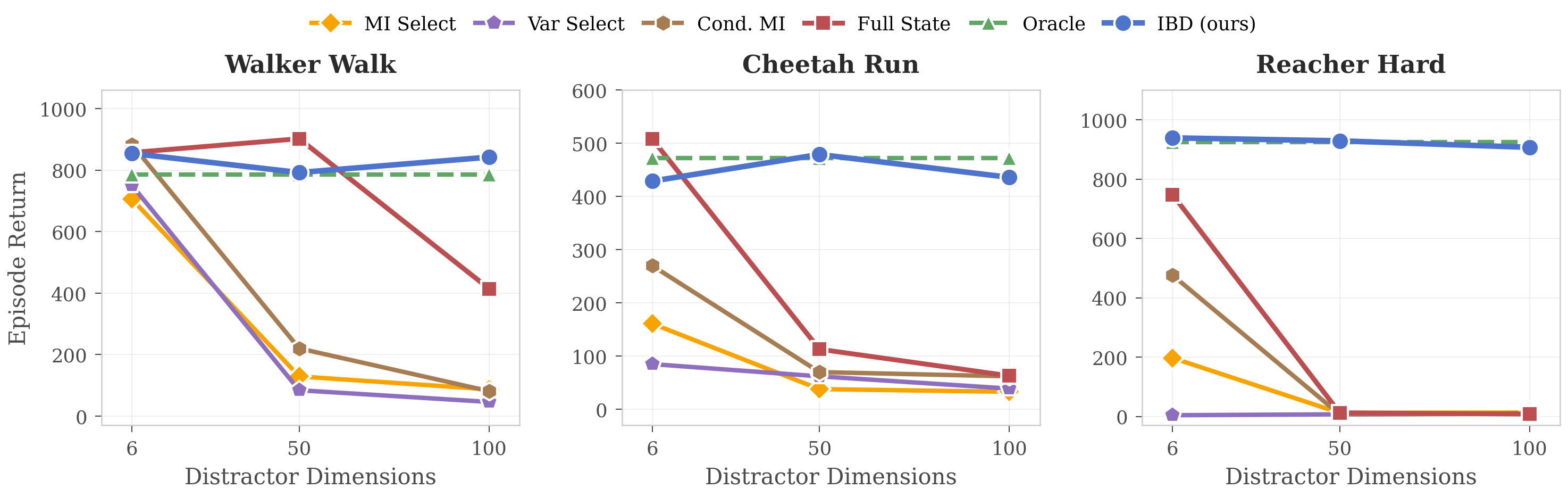}
\caption{\textbf{Distractor scaling curve.}
Episode return as a function of distractor dimensionality
(6, 50, 100) for \texttt{walker\_walk} (left),
\texttt{cheetah\_run} (center), and \texttt{reacher\_hard} (right).
Full State (red) degrades as distractors increase;
\IBD{} (blue) tracks oracle performance across all
distractor counts.
Shaded regions: $\pm$1 std over 5 seeds.
}
\label{fig:scaling}
\end{figure}

\vspace{-4mm}
\subsection{Scaling with Distractor Dimension}
\label{sec:scaling}
\vspace{-3mm}
As shown in Figure~\ref{fig:scaling}, we measured how return changes as the number of distractors increases from 6 to 50 to 100. Full-State SAC remains competitive when the distractor-to-signal ratio is small, but degrades once distractors dominate the observation vector. The failure point is better predicted by this ratio than by the absolute distractor count. For \texttt{walker\_walk}, which has 24 true dimensions, Full State tolerates 50 distractors but drops sharply at 100. For \texttt{reacher\_hard}, which has only 6 true dimensions, Full State collapses already at 50 distractors.

\IBD{} remains close to Oracle across the scaling curve because the per-dimension test asks whether a coordinate responds to action intervention, not whether it resembles other coordinates passively. A denser five-point sweep ($d_d\in\{6,24,50,100,150\}$, $3$ seeds per cell) confirms this on the distractor-to-signal ratio axis: in the degradation regime $d_d\geq d_c$, a descriptive log--log fit gives $R/R_{\mathrm{oracle}}\propto(d_d/d_c)^{-1.71}$ for Full State, while \IBD{}'s exponent is $0.002$, consistent with no degradation. Fit details, confidence intervals, and per-task curves are in Appendix~\ref{app:scaling_law}.

\vspace{-4mm}
\subsection{Causal Boundary Recovery}
\vspace{-3mm}

We next test whether the downstream gains correspond to accurate recovery of the action-reachable boundary. Across all $12$ settings, \IBD{} achieves mean precision, recall, and F1 of $0.95$, with precision at least $0.93$ in $11$ of the $12$ settings, so FDR-controlled intervention testing rarely readmits the exogenous distractors that feature selection is meant to remove. Recall drops only where some true dimensions have a weak intervention response (\texttt{walker\_walk} $0.85$--$0.92$, \texttt{hopper\_hop} $0.75$), and downstream return stays close to Oracle in those cases, suggesting that the missed dimensions are redundant with selected state variables or weakly tied to the reward. Boundary quality is also stable as distractor count grows: on \texttt{cheetah\_run}, F1 is $0.99$, $0.97$, and $0.98$ from easy to medium to hard. Per-setting numbers are in \Cref{tab:boundary} (Appendix~\ref{app:boundary}); the mask-level results agree with the control results, indicating that \IBD{} improves return by excluding non-causal dimensions rather than by overfitting a specific learner.

\vspace{-4mm}
\subsection{Stronger Observational Dynamics Baselines}
\label{sec:obs_dyn}
\vspace{-3mm}
We further compare against two model-based observational baselines. The \emph{gradient-attribution} baseline trains a joint forward model $f(\vs,\va)\to\vs'$ and ranks each output dimension by $\E[\|\partial f_i/\partial\va\|_1]$. The \emph{multistep inverse dynamics} baseline adapts the action-recovery signal used in exogenous-block MDP methods~\citep{efroni2022provably,levine2025exbmdp}: it trains $g(\vo_t,\vo_{t+k})\to\va_t$ and ranks dimensions by the validation MSE increase caused by ablating each dimension. Both baselines use natural action variation in collected rollouts rather than action interventions. (Implementation details are given in Appendix~\ref{app:grad_attr} and~\ref{app:inverse_dyn}.)

\begin{table}[t]
\centering
\small
\caption{Multistep inverse dynamics baseline: episode return (mean $\pm$ std over 5 seeds, medium/hard distractors) and boundary discovery accuracy (precision / recall / F1). \textbf{Bold:} best non-oracle method on return.}
\label{tab:inverse_dyn}
\setlength{\tabcolsep}{4pt}
\begin{tabular}{ll c cc cc c}
\toprule
&& & \multicolumn{2}{c}{Inverse Dyn.} & \multicolumn{2}{c}{\IBD{} (ours)} & \\
\cmidrule(lr){4-5} \cmidrule(lr){6-7}
Environment & Distr. & Full State & Return & P / R / F1 & Return & P / R / F1 & Oracle \\
\midrule
\texttt{reacher\_hard} & medium & 12$\pm$7    & 28$\pm$53  & 0.67 / 0.67 / 0.67 & \textbf{929}$\pm$76  & \textbf{0.97 / 1.00 / 0.98} & 925$\pm$75  \\
\texttt{walker\_walk}  & hard   & 414$\pm$144 & 624$\pm$26 & 0.88 / 0.88 / 0.88 & \textbf{842}$\pm$121 & \textbf{0.97 / 0.85 / 0.91} & 785$\pm$128 \\
\texttt{cheetah\_run}  & medium & 113$\pm$34  & 285$\pm$74 & 0.95 / 0.95 / 0.95 & \textbf{479}$\pm$95  & \textbf{0.94 / 1.00 / 0.97} & 472$\pm$35  \\
\texttt{finger\_spin}  & medium & 365$\pm$13  & 21$\pm$26  & 0.44 / 0.44 / 0.44 & \textbf{587}$\pm$178 & \textbf{0.96 / 1.00 / 0.98} & 775$\pm$181 \\
\bottomrule
\end{tabular}
\end{table}

Neither model-based observational baseline closes the gap. In Table~\ref{tab:grad_attr}, gradient attribution reaches only $5\pm10$ on \texttt{reacher\_hard}, compared with $929\pm76$ for \IBD{}, and reaches $56\pm32$ on \texttt{cartpole\_swingup}, below the simpler passive selectors in Table~\ref{tab:main}. Its boundary F1 is only $0.44$--$0.50$, meaning that roughly half of the selected dimensions are distractors. A learned forward model can assign action sensitivity to mimicking distractors when they co-vary with action-influenced state under the observational rollout distribution.

Multistep inverse dynamics is stronger than raw MI or variance and improves over Full State in $3$ of $4$ settings in Table~\ref{tab:inverse_dyn}. However, it still does not match \IBD{} or Oracle. On \texttt{reacher\_hard} medium, inverse dynamics obtains F1 $0.67$, so two of the six selected dimensions are distractors; this small leakage reduces return from $929$ to $28$. On \texttt{finger\_spin} medium, F1 falls to $0.44$ and return drops below Full State. These results show that action-recovery objectives can still confuse causal influence with passive co-variation unless the action mechanism is intervened on.

\vspace{-4mm}
\subsection{Confounded Distractors}
\label{sec:confounding}
\vspace{-3mm}

\begin{table}[t]
\centering
\small
\caption{\textbf{Confounded distractor regime} (boundary F1, mean over \texttt{walker\_walk}, \texttt{cheetah\_run}, and \texttt{reacher\_hard}, $5$ seeds each). A latent AR(1) process $C_t$ enters the action channel with strength $\kappa$ and drives $12$ distractor dimensions with strength $\lambda$; the backdoor path is open only when both edges are active ($\kappa{=}0.5,\lambda{=}1$). Per-task results and false-positive breakdowns are in Appendix~\ref{app:confounding}.}
\label{tab:confounding}
\begin{tabular}{lcccc}
\toprule
Method & $\kappa{=}0,\lambda{=}0$ & $\kappa{=}0.5,\lambda{=}0$ & $\kappa{=}0,\lambda{=}1$ & $\kappa{=}0.5,\lambda{=}1$ \\
\midrule
\IBD{} (ours)  & \textbf{0.934} & \textbf{0.934} & \textbf{0.949} & \textbf{0.949} \\
Inverse Dyn.   & 0.870 & 0.833 & 0.872 & 0.733 \\
MI Select      & 0.919 & 0.724 & 0.919 & 0.403 \\
Grad.\ Attr.   & 0.563 & 0.531 & 0.559 & 0.457 \\
Cond.\ MI      & 0.498 & 0.456 & 0.499 & 0.455 \\
Var Select     & 0.329 & 0.356 & 0.359 & 0.368 \\
\bottomrule
\end{tabular}
\vspace{-4mm}
\end{table}

The distractors used so far are appended exogenous processes, so Proposition~\ref{prop:obs_fail} is illustrated by construction rather than instantiated in the benchmark. We therefore add a regime in which a genuine latent confounder enters the action channel, realizing the backdoor path of Proposition~\ref{prop:obs_fail} in a continuous-control environment. A latent AR(1) process $C_t$ perturbs the probe policy's actions with strength $\kappa$ and independently drives $12$ appended distractor dimensions with strength $\lambda$, with no edge from any action or state to those dimensions. The $2\times2$ design over $(\kappa,\lambda)$ isolates the effect of opening the backdoor path: the ground-truth action-reachable set is identical in all four cells, so F1 is comparable across the columns of \Cref{tab:confounding}. Details of the generative process are in Appendix~\ref{app:confounding}.

The control cell $\kappa{=}\lambda{=}0$ confirms that the observational baselines are not simply mis-implemented: MI Select reaches F1 $0.919$ there, and multistep inverse dynamics $0.870$. Opening a single edge does not create a backdoor path, since $C_t$ then influences either the actions or the distractors but not both. Only when both edges are active is the backdoor path open, and the observational selectors degrade sharply: MI Select falls to $0.403$, admitting $8.5$ of the $12$ confounded dimensions, while inverse dynamics falls to $0.733$ with $3.4$ confounded false positives. \IBD{} is essentially unchanged across all four cells (F1 $0.934$--$0.949$, identical for $\kappa{=}0$ and $\kappa{=}0.5$), because randomizing the action mechanism severs $C \to \va$ while leaving the autonomous mechanism of the distractors intact, exactly as Proposition~\ref{prop:confound} predicts.

\vspace{-4mm}
\subsection{Transfer across Downstream RL Algorithms}
\label{sec:td3}
\vspace{-3mm}



To test whether the benefit depends on SAC, we repeat the hard-distractor experiments on \texttt{walker\_walk} and \texttt{cheetah\_run} with TD3~\citep{fujimoto2018td3} as the downstream algorithm. The same mask transfers without change. On \texttt{walker\_walk}, TD3 with Full State reaches $142\pm131$, while TD3 with \IBD{} reaches $734\pm208$ against Oracle at $781\pm116$. On \texttt{cheetah\_run}, Full State reaches $67\pm12$, while \IBD{} reaches $356\pm80$ against Oracle at $425\pm24$. Thus, \IBD{} reaches $94\%$ and $84\%$ of oracle return on the two tasks.

The gain is larger for TD3 than for SAC on \texttt{walker\_walk}: \IBD{} improves TD3 Full State by more than $5\times$, compared with about $2\times$ for SAC. This is consistent with TD3 being more sensitive to irrelevant high-dimensional inputs because it lacks SAC's entropy-regularized training objective. Since the mask is computed before downstream RL training and is unchanged across learners, the result indicates that the main benefit comes from causal feature filtering rather than from a SAC-specific effect.

\begin{figure}[t]
\centering
\includegraphics[width=\textwidth]{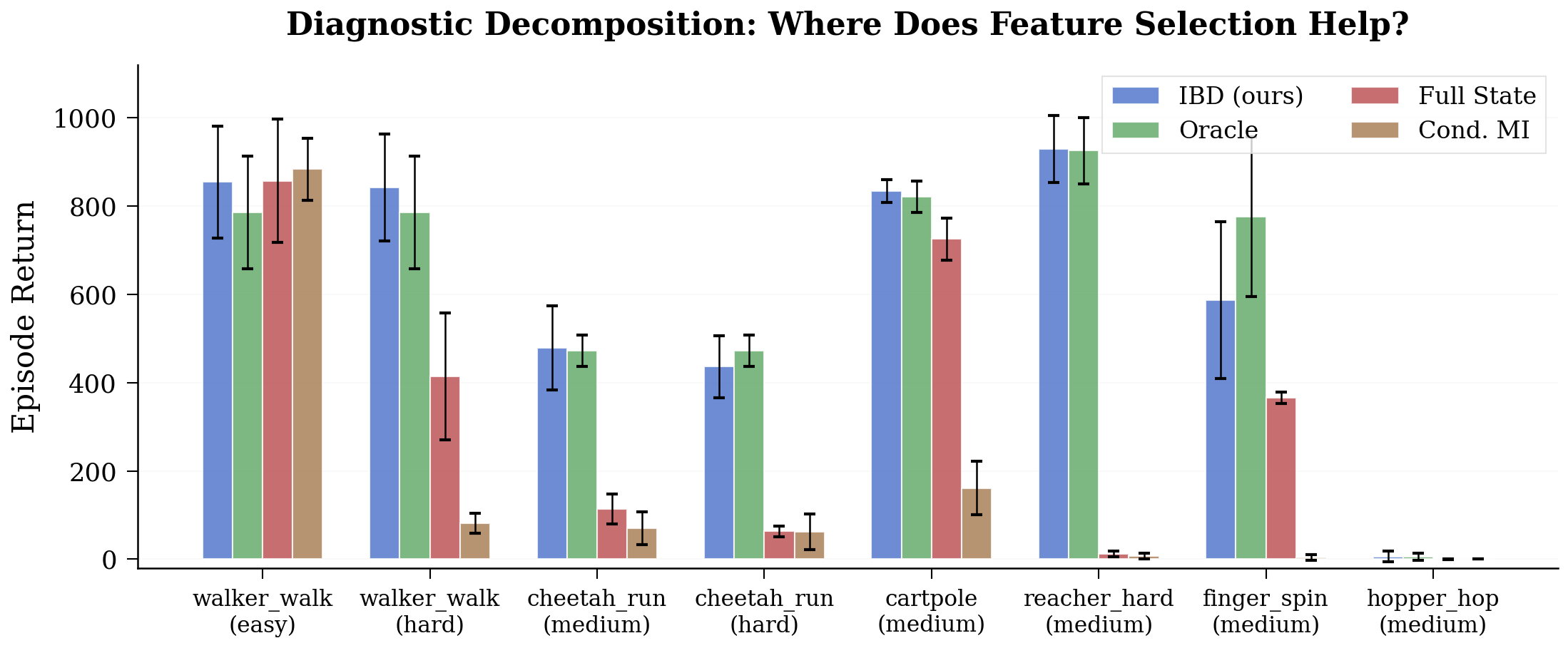}
\caption{\textbf{Where does feature selection help?}
Episode return across 8 representative settings.
Three regimes emerge. (1)~\texttt{walker\_walk} (easy): all methods perform similarly, so no selection is needed. (2)~Most medium/hard settings: Oracle $\gg$ Full State but \IBD{} $\approx$ Oracle, so distractors are the bottleneck and \IBD{} resolves them. (3)~\texttt{hopper\_hop}: all methods near zero, so the bottleneck is exploration, not feature selection.}
\label{fig:diagnostic}
\end{figure}

\vspace{-4mm}
\subsection{Partially Controllable Dimensions}
\label{sec:robustness}
\vspace{-3mm}


In the real system, dimensions might contain both action-dependent and autonomous components. Thus, we test this setting by adding partially controllable dimensions with dynamics
\begin{equation}
\label{eq:partial}
x_{t+1} = \alpha \cdot g(\vs_t, \va_t) + (1 - \alpha) \cdot z_t,
\end{equation}
where $g$ is a nonlinear function of state and action, $z_t$ is an exogenous OU process, and $\alpha\in[0,1]$ controls the strength of causal mixing.

\IBD{} detects weak but real action influence. At $\alpha=0$, the added dimensions are purely exogenous and recall is correctly zero. By $\alpha\approx0.05$, corresponding to only about $5\%$ causal variance, recall reaches $1.0$ on both \texttt{cheetah\_run} and \texttt{walker\_walk}. Precision remains at least $0.92$ across all tested values of $\alpha$. The intervention test therefore behaves as a graded detector of action influence rather than as a brittle binary test.
\vspace{-4mm}
\subsection{Diagnostic Decomposition}
\vspace{-3mm}


Figure~\ref{fig:diagnostic} turns the comparison among Full State, \IBD{}, and Oracle into a diagnostic. The three regimes it identifies map onto distinct remedies: when Full State already matches Oracle no selection is needed; when Oracle dominates and \IBD{} closes the gap, as in \texttt{reacher\_hard} medium and \texttt{cheetah\_run} medium/hard, the bottleneck is distractor confusion; and when all three fail the problem lies in exploration or reward design rather than in the observation space.

\vspace{-4mm}
\section{Discussion}
\label{sec:discussion}
\vspace{-3mm}

The degradation reflects a finite-capacity RL failure under high-dimensional irrelevant inputs, not a changed control problem: distractors do not affect the true state, reward, actions, or transitions.
Although sufficiently large policies could learn to ignore them, finite-budget training can overvalue non-causal dimensions correlated with state statistics or episode progress.
\IBD{} addresses this before policy learning by applying an interventional observation mask, and can also operate on learned encoder features~\citep{yarats2022drqv2,laskin2020curl}.
Our setting assumes stationary probing with appended exogenous distractors; indirect action effects, policy-dependent distractors, and different capacity--budget regimes remain future directions~\citep{efroni2022provably,levine2025exbmdp}.

\begin{ack}
The authors J.L., A.C., and P.B. acknowledge the support by the National Science Foundation (NSF) Award 2243104 under the Center for Complex Particle Systems (COMPASS), the Career Award CPS-1453860, CNS-1932620, and NSF Mid-Career Advancement Award BCS-2527046, the U.S. Army Research Office (ARO) under Grant No. W911NF-23-1-0111, the Defense Advanced Research Projects Agency (DARPA) Young Faculty Award and DARPA Director Fellowship Award under Grant Number N66001-17-1-4044, Intel faculty awards and a Northrop Grumman grant. P.B. is grateful to National Institute of Health (NIH) for the grants R01 AG 079957 ``Interpretable machine learning to synergize brain age estimation and neuroimaging genetics'' and RF1 AG 082201 ``Neurovascular calcification and ADRD in two nonindustrial Native American populations''. P.B. is also grateful to Carnegie Mellon University for hosting him during his sabbatical. The views, opinions, and/or findings in this article are those of the authors and should not be interpreted as official views or policies of the Department of War, the National Institute of Health or the National Science Foundation.
\end{ack}


\bibliographystyle{unsrtnat}
\bibliography{references}

\newpage
\clearpage
\appendix

\section{Distractor Environment Design}
\label{app:distractors}

Our distractor benchmark instantiates three distractor patterns that recur in instrumented physical and software-driven systems. Each captures a different statistical regime that an observation-based selector would face in deployment.

\emph{Autonomous distractors} are Ornstein-Uhlenbeck processes and coupled oscillators that evolve independently of the agent's actions and state. They model environmental noise sources (vibration, ambient processes, secondary equipment cycles) whose variance and frequency content differ from true state dimensions, and are filterable by any method that examines statistical properties beyond raw magnitude.

\emph{Mimicking distractors} are exogenous OU processes whose marginal statistics are matched to those of true proprioceptive dimensions, modeling sensor cross-talk and time-correlated readout artifacts that share statistical signatures with the channels they sit alongside. The match targets scale, variance, autocorrelation, and frequency content, using $\sigma_{\mathrm{ref}} = 0.5 + 0.05 \cdot d_c$ so that each task receives appropriately scaled distractors (\texttt{walker}: $\sigma_{\mathrm{ref}} = 1.7$; \texttt{cartpole}: $\sigma_{\mathrm{ref}} = 0.75$). Because they are indistinguishable from true dimensions in their marginal distribution, methods that rely solely on marginal statistics cannot separate them from genuine state.

\emph{Reward-correlated distractors} are autonomous processes whose temporal profile resembles the within-episode reward curve typical of locomotion tasks: a saturating ramp $1 - \exp(-\kappa t)$, gait-frequency sinusoidal oscillation, and OU noise. Their dynamics depend only on episode time $t$, not on state, reward, or actions; the resemblance to reward curves nonetheless induces spurious mutual information with the agent's action sequence (because policy actions also accumulate over the episode), even though no causal path from actions to these dimensions exists.

The composition at each difficulty level is as follows. Easy (6 total): 4~OU + 2~oscillator. Medium (50 total): 12~autonomous + 28~mimicking + 6~reward-correlated + 4~oscillator. Hard (100 total): 18~autonomous + 60~mimicking + 14~reward-correlated + 8~oscillator. Distinguishing all three types requires reasoning about causal influence rather than statistical association, since an action intervention leaves the distribution of every type unchanged regardless of its marginal similarity to true state.

\section{Partial Controllability: Experimental Details}
\label{app:robustness}

\begin{figure}[h]
\centering
\includegraphics[width=0.6\textwidth]{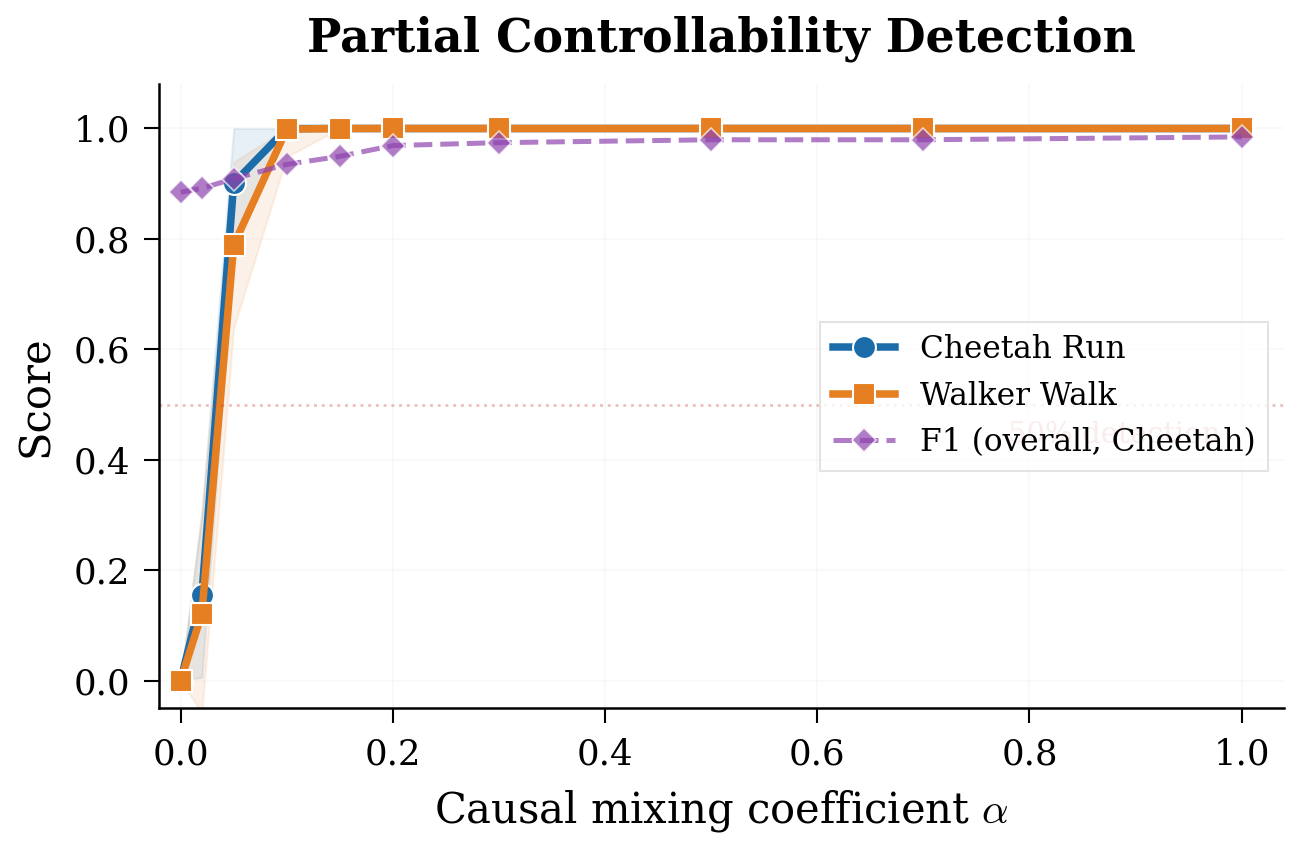}
\caption{\textbf{Partial controllability detection.}
Recall on partially controllable dimensions as a function of mixing coefficient $\alpha$, for \texttt{cheetah\_run} and \texttt{walker\_walk}.
At $\alpha = 0$ (purely exogenous), recall is correctly zero; by $\alpha \approx 0.05$ (${\sim}5\%$ causal variance), recall reaches 1.0.
The dashed line shows overall F1 on \texttt{cheetah\_run}, confirming that boundary quality remains high.
Precision $\geq 0.92$ at all $\alpha$ values.
Shaded regions: $\pm$1 std over 3 seeds.
}
\label{fig:robustness}
\end{figure}

For the robustness study (\Cref{sec:robustness}), each partially controllable dimension follows the dynamics in \Cref{eq:partial}, where the causal component $g$ maps a randomly-assigned action dimension through a nonlinear transformation (tanh with per-dimension gain and bias), integrated via leaky dynamics with rate constant $50 \times dt$ to ensure the causal signal is on the same scale as the exogenous OU noise.
The exogenous component $z_t$ is an independent OU process with $\tau \in [1, 4]$ and $\sigma \in [0.2, 0.6]$, matching the scale of DMControl proprioceptive dimensions.

Each configuration adds 6 partially controllable dimensions and 20 exogenous distractors (10 autonomous OU + 10 mimicking) to the base DMControl task.
We sweep 10 values of $\alpha$ across 3 seeds per task, for a total of 60 IBD probes.
Each probe takes approximately 2--3 minutes (trajectory collection: 160 trajectories $\times$ 200 steps; statistical testing: $<1$s).

\section{Scout Policy Ablation}
\label{app:scout_ablation}

We ablate the scout training budget across $\{0, 10\text{K}, 20\text{K}, 40\text{K}, 80\text{K}, 160\text{K}\}$ steps on three tasks with medium distractors, using 3 seeds per configuration.
Budget $= 0$ corresponds to \IBD{}'s structured random probe policy, which requires no RL training and no GPU.

\begin{table}[h]
\centering
\caption{Scout ablation (P/R/F1, mean over 3 seeds). Budget $= 0$: structured random probe; $\geq$10K: SAC scout. Boundary accuracy is identical or better without RL pre-training.}
\label{tab:scout_ablation}
\begin{tabular}{llccc}
\toprule
Task & Scout Budget & Precision & Recall & F1 \\
\midrule
\texttt{cheetah\_run}  & 0    & 0.944 & 1.000 & 0.971 \\
                        & 80K  & 0.944 & 1.000 & 0.971 \\
\midrule
\texttt{reacher\_hard}  & 0   & 0.952 & 1.000 & 0.974 \\
                         & 80K & 0.952 & 1.000 & 0.974 \\
\midrule
\texttt{walker\_walk}   & 0   & 0.910 & 0.972 & 0.940 \\
                         & 80K & 0.902 & 0.889 & 0.895 \\
\bottomrule
\end{tabular}
\end{table}

On \texttt{cheetah\_run} and \texttt{reacher\_hard}, boundary accuracy is identical across all budgets; intermediate values all match budget $= 0$ and are omitted.
On \texttt{walker\_walk}, the untrained policy achieves the highest F1 (0.94 versus 0.90 at 80K), likely because a trained policy converges toward a narrow behavioral mode that reduces state coverage.

\section{Boundary Discovery Accuracy: Full Results}
\label{app:boundary}

\input{table3_boundary}

\Cref{tab:boundary} reports \IBD{}'s per-setting precision, recall, and F1.
Precision is at least $0.93$ in $11$ of $12$ settings and $0.86$ on \texttt{reacher\_hard} hard, confirming that FDR control keeps distractor leakage low.
Recall reaches 1.00 on most tasks; the two exceptions are \texttt{walker\_walk} (0.85--0.92, where some proprioceptive dimensions have weak interventional signal) and \texttt{hopper\_hop} (0.75, an intrinsically difficult exploration task).
Boundary accuracy is stable across distractor counts: for \texttt{reacher\_hard}, F1 is 0.97 (easy), 0.98 (medium), and 0.93 (hard).

\section{Hyperparameters}
\label{app:hparams}

\begin{table}[H]
\centering
\caption{Hyperparameters used in all experiments.}
\label{tab:hparams}
\resizebox{\linewidth}{!}{%
\begin{tabular}{lll}
\toprule
Component & Parameter & Value \\
\midrule
SAC / TD3 & LR / batch / buffer / arch & $3{\times}10^{-4}$ / 256 / 300K / MLP [256, 256] \\
TD3 only & Action noise $\sigma$ & 0.1 \\
\midrule
\IBD{} Probe & Trajectories $N$ / length $T$ / horizons $\cH$ / $\alpha$ & 80 / 200 / \{1,5,10\} / 0.05 \\
\midrule
Training & Steps / eval freq / eval eps / seeds & 300K / 50K / 10 / \{42,142,242,342,442\} \\
\midrule
Cond.\ MI & Arch / epochs / LR / batch / data & MLP [64,64] / 50 / $10^{-3}$ / 2048 / 200$\times$200 \\
Grad.\ Attr. & Arch / epochs / LR / batch / data & MLP [128,128] / 80 / $10^{-3}$ / 2048 / 200$\times$200 \\
Inverse Dyn. & Arch / epochs / LR / batch / data / horizon $k$ & MLP [128,128] / 80 / $10^{-3}$ / 2048 / 200$\times$200 / 3 \\
\midrule
Robustness & $\alpha$ / partial dims / exo distr / seeds & \{0\ldots1.0\} / 6 / 20 / \{42,142,242\} \\
\bottomrule
\end{tabular}}
\end{table}

\paragraph{Compute resources.}
All experiments run on CPU; no GPU is required. The \IBD{} probing step uses only NumPy and SciPy (Welch's $t$-test, Benjamini--Hochberg correction) and contains no neural network: a single probe (32K environment steps with $N{=}80$, $T{=}200$) completes in 2--3 minutes on one CPU core (Appendix~\ref{app:robustness}). Downstream SAC/TD3 training uses Stable-Baselines3 with the default CPU device, which is faster than GPU for the MLP [256, 256] policies due to data-transfer overhead; one 300K-step training run takes approximately 27 minutes on a single CPU core (measured across 99 runs spanning all six tasks: min 25.6, max 29.3, mean 27.0 minutes).

Total compute for the experiments reported in this paper is approximately 280 CPU-hours, broken down as: \Cref{tab:main} main results (12 settings $\times$ 6 methods $\times$ 5 seeds $\approx 360$ runs, ${\sim}160$ CPU-h); the TD3 validation of \Cref{sec:td3} (${\sim}14$ CPU-h); \Cref{tab:grad_attr,tab:inverse_dyn} model-based baselines (${\sim}18$ CPU-h); the dense scaling sweep in Appendix~\ref{app:scaling_law} (${\sim}60$ CPU-h); the partial-controllability and scout-budget ablations (${\sim}6$ CPU-h); and the confounded-distractor, summary-statistic, and multiplicity studies of Appendices~\ref{app:confounding}--\ref{app:multiplicity} (${\sim}25$ CPU-h), which are mask-level only: no downstream policy is trained and each probe costs $2$--$3$ minutes.

\section{Gradient Attribution Baseline: Implementation Details}
\label{app:grad_attr}

The gradient attribution baseline (results in \Cref{tab:grad_attr}) trains a joint forward dynamics model $f(\vs, \va) \to \vs'$ (MLP [128, 128], 80 epochs) and scores each observation dimension $i$ by the mean absolute gradient of the predicted next state with respect to actions:
$\text{score}_i = \mathbb{E}_{\text{data}}\!\big[\| \partial f_i(\vs, \va) / \partial \va \|_1 \big]$.
This directly measures how sensitive the learned dynamics are to the action input and represents a direct model-based approach to identifying action-influenced dimensions.

However, gradient attribution remains observational. When mimicking distractors co-vary with true state dimensions that also co-vary with actions, a learned forward model can assign action sensitivity to those distractors under the observational data distribution, even though the interventional effect of $\doop(\va)$ on them is zero. Its boundary discovery F1 is accordingly low (0.44--0.50), indicating that gradient attribution selects roughly half distractors and half true dimensions.

\section{Multistep Inverse Dynamics Baseline: Implementation Details}
\label{app:inverse_dyn}

As a further test of whether observational methods with stronger structural priors can succeed where MI, Cond.\ MI, and gradient attribution fail, we evaluate a continuous-control adaptation of the multistep inverse dynamics principle from the EX-BMDP line~\citep{efroni2022provably,levine2025exbmdp,levine2025craft} (results in \Cref{tab:inverse_dyn}).
We do not compare directly against STEEL~\citep{levine2025exbmdp}, whose discrete latent state recovery, fast-mixing assumption, and PAC-style $|S|$ upper bound do not transfer to continuous state-action spaces without substantial reformulation.
Instead, we isolate the multistep inverse dynamics signal that is the core empirical ingredient of that line and apply it at the observation-dimension level, matching our budget-controlled feature selection protocol.

The method trains a network $g(\vo_t, \vo_{t+k}) \to \va_t$ (MLP [128, 128], 80 epochs) on rollouts collected with the same structured-random probe policy used by \IBD{}, with horizon $k = 3$.
For each observation dimension $i$, we score dimension $i$ by the increase in validation MSE when dimension $i$ is zeroed out in both $\vo_t$ and $\vo_{t+k}$ at test time: $\mathrm{score}_i = \mathrm{MSE}_{\mathrm{ablated}}(i) - \mathrm{MSE}_{\mathrm{full}}$.
Dimensions whose values carry action information receive high scores; purely exogenous distractors should receive low scores.
The top-$d_c$ dimensions are selected, matching the budget given to MI, Variance, and Cond.\ MI.

This baseline remains observational: it exploits natural action variation in the data collection policy, not interventions. It is therefore still vulnerable to confounded distractors whose dynamics co-vary with action-influenced state dimensions (\Cref{sec:confounding}), in contrast to \IBD{}, which severs such confounding by replacing the action mechanism with a confounder-independent randomized policy.

\section{Additional Distractor Scaling Sweep}
\label{app:scaling_law}

This appendix supplements \Cref{sec:scaling} with a denser distractor sweep on the distractor-to-signal ratio axis $d / d_c$. Across this sweep, Full State return decreases as $d / d_c$ grows, while \IBD{} stays close to Oracle. We summarize the trend with a power-law fit; the fit is descriptive on the points we sweep, not a claim about a universal scaling exponent.

\paragraph{Setup.}
We sweep five distractor counts, $d \in \{6, 24, 50, 100, 150\}$, on \texttt{walker\_walk} ($d_c = 24$), \texttt{cheetah\_run} ($d_c = 17$), and \texttt{reacher\_hard} ($d_c = 6$), giving distractor-to-signal ratios from $0.25$ up to $25$. Three downstream RL methods are evaluated: Full State (raw observation), \IBD{} (interventional mask), and Oracle (ground-truth action-reachable dimensions). All hyperparameters match \Cref{sec:setup}: SAC, $300$K environment steps, MLP $[256, 256]$, $3$ seeds per cell. Oracle is mask-invariant in distractor count and reported once per task.

\begin{figure}[t]
\centering
\includegraphics[width=0.95\textwidth]{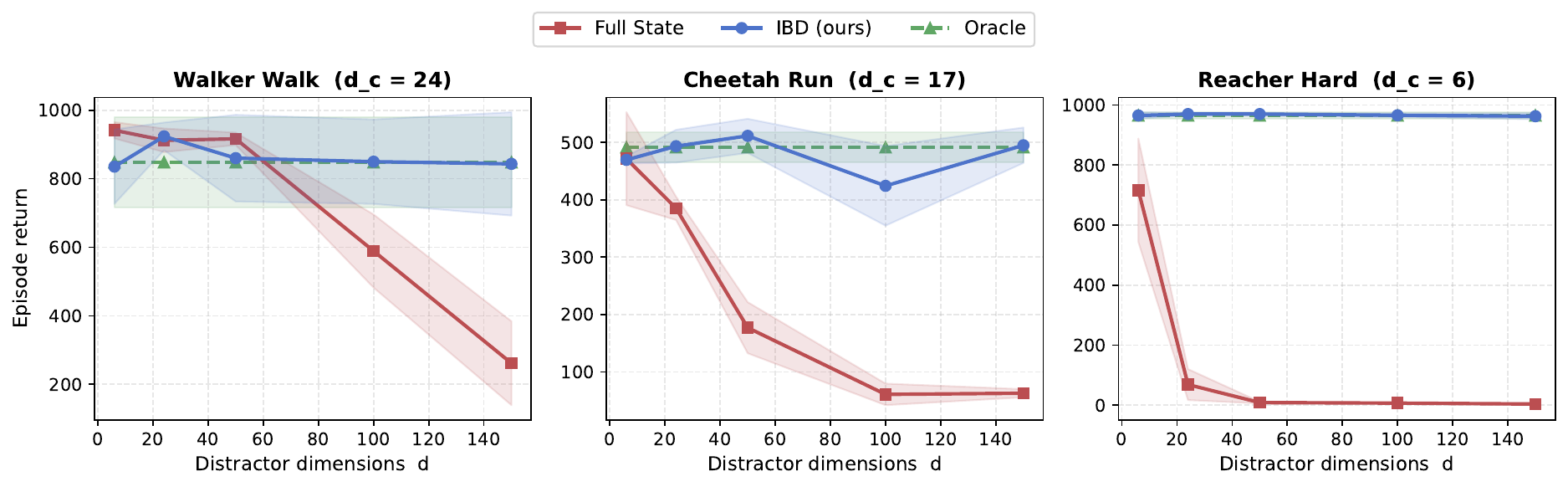}
\caption{\textbf{Distractor scaling curves (dense sweep).} Episode return vs.\ distractor count $d$ for \texttt{walker\_walk} ($d_c\!=\!24$, left), \texttt{cheetah\_run} ($d_c\!=\!17$, center), and \texttt{reacher\_hard} ($d_c\!=\!6$, right). Full State (red) collapses earlier on tasks with smaller $d_c$; \IBD{} (blue) tracks Oracle across all distractor counts. Shaded regions: $\pm 1$ std over $3$ seeds.}
\label{fig:scaling_curves_dense}
\end{figure}

\paragraph{Linear-scale curves.}
\Cref{fig:scaling_curves_dense} shows episode return as a function of distractor count for each task. Three failure profiles emerge. \texttt{Walker\_walk} ($d_c = 24$) tolerates distractors up to $d = 50$ ($d/d_c \approx 2$) and only collapses at $d = 100$. \texttt{Cheetah\_run} ($d_c = 17$) degrades monotonically from $d = 24$ onward. \texttt{Reacher\_hard} ($d_c = 6$) collapses already at $d = 24$ ($d/d_c = 4$) and stays near zero return through $d = 150$. \IBD{} tracks Oracle in all three tasks across the full sweep.

\begin{figure}[t]
\centering
\includegraphics[width=0.78\textwidth]{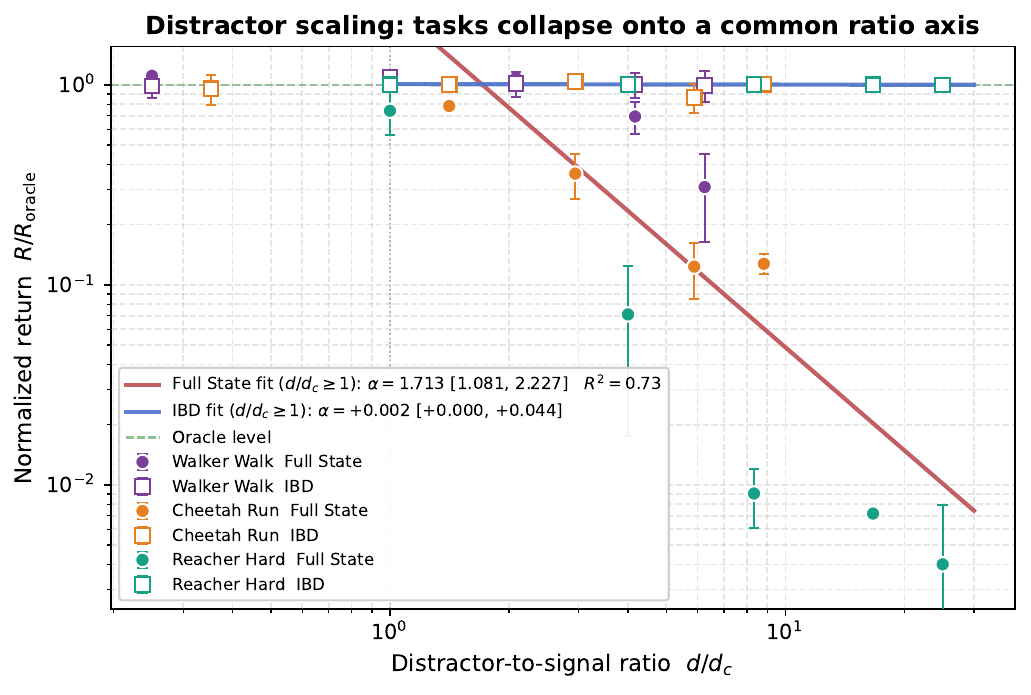}
\caption{\textbf{Distractor scaling on the ratio axis.} Normalized return $R / R_{\mathrm{oracle}}$ vs.\ distractor-to-signal ratio $d / d_c$ on log--log axes. Filled markers: Full State; open markers: \IBD{}; colors distinguish tasks. Solid lines are descriptive power-law fits restricted to $d \geq d_c$ (vertical dotted line). Pooled across tasks, Full State follows $R/R_{\mathrm{oracle}} \propto (d/d_c)^{-1.71}$ ($R^2 = 0.73$); \IBD{} stays at the oracle level with $\alpha \approx 0$.}
\label{fig:scaling_law}
\end{figure}

\paragraph{Power-law fit on the ratio axis.}
\Cref{fig:scaling_law} plots normalized return $R / R_{\mathrm{oracle}}$ against the ratio $d / d_c$ on log--log axes. We fit a power law $R / R_{\mathrm{oracle}} = a \cdot (d / d_c)^{-\alpha}$ on the degradation regime $d \geq d_c$, since for $d < d_c$ distractors are a minority of observation dimensions and the network has sufficient capacity to ignore them, so a power-law model is not appropriate there. Fits are done by weighted least squares in log space, with $95\%$ confidence intervals on $\alpha$ from $1000$ bootstrap resamples.

Pooling Full State across the three tasks ($n = 13$ ratio--task points), we obtain
\[
\alpha_{\mathrm{Full\,State}} = 1.71 \quad [1.08,\, 2.23], \qquad R^2 = 0.73,
\]
and pooling \IBD{} on the same set,
\[
\alpha_{\IBD{}} = 0.002 \quad [0.000,\, 0.044].
\]
The two bootstrap CIs do not overlap on this sweep: \IBD{}'s exponent is consistent with no degradation across nearly two decades of $d / d_c$, while Full State decays roughly as $(d / d_c)^{-1.7}$ on the swept range.

\paragraph{Connection to the collapse threshold.}
The pooled exponent $\alpha \approx 1.71$ is consistent with the per-task collapse points implied by the curves in \Cref{fig:scaling_curves_dense}: Full State return drops below $50\%$ of its easy-distractor baseline at distractor-to-signal ratios of $4.2{:}1$ (\texttt{walker\_walk}), $2.9{:}1$ (\texttt{cheetah\_run}), and $8.3{:}1$ (\texttt{reacher\_hard}). A power-law extrapolation with slope $\alpha \approx 1.71$ predicts that $R / R_{\mathrm{oracle}}$ falls below $0.5$ near $d / d_c \approx 1.7$, on the same order as these observed ratios. The variation across tasks reflects task-specific reward shape and policy robustness around the collapse. \IBD{}'s near-zero exponent on the same axis quantifies the qualitative claim of \Cref{sec:scaling} that interventional masking decouples downstream return from the distractor-to-signal ratio.

\section{Confounded Distractor Regime: Design and Per-Task Results}
\label{app:confounding}

This appendix details the confounded regime of \Cref{sec:confounding}. A latent process, never observed by any method, follows a discrete-time AR(1) recursion with unit stationary variance,
\[
C_t = \rho_c \, C_{t-1} + \sqrt{1 - \rho_c^2}\; \xi_t, \qquad \xi_t \sim \mathcal{N}(0, I),
\]
and enters the system through two edges whose strengths are swept independently. The probe policy's action is perturbed by $C_t$ with strength $\kappa$,
\[
\va_t = \mathrm{clip}\big( \pi_{\mathrm{probe}}(\vo_t) + \kappa \, K C_t \big),
\]
and the confounded distractor block is driven by the same latent process with strength $\lambda$,
\[
\vd_t = \rho \, \vd_{t-1} + \lambda \, L C_t + \varepsilon_t .
\]
There is no edge from $\va_t$ or $\vs_t$ into $\vd_t$, so the confounded dimensions remain outside the SoI by construction and the oracle mask (true state dimensions only) stays valid in all four cells. When both $\kappa > 0$ and $\lambda > 0$, however, $\va_t$ and $\vd_t$ share a latent cause, which opens the backdoor path of Proposition~\ref{prop:obs_fail}: $I(\vd_{t+1}; \va_t) > 0$, a state-conditioned forward model gains predictive power from $\va_t$ because actions reveal $C_t$ that $\vs_t$ does not determine, and $\va_t$ becomes partially recoverable from observation pairs that include $\vd$. Under $\doop(\va \sim \mathrm{Unif}(\cA))$ the edge $C \to \va$ is severed while the mechanism generating $\vd$ is untouched, so \IBD{}'s two-sample test sees no shift on those dimensions.

Each task receives $12$ confounded dimensions in addition to the medium exogenous distractor bank, giving observation dimensions $74$ (\texttt{walker\_walk}), $67$ (\texttt{cheetah\_run}), and $56$ (\texttt{reacher\_hard}). We use $\rho_c = 0.95$ for the latent process and sweep $\kappa \in \{0, 0.5\}$, $\lambda \in \{0, 1\}$ with $5$ seeds per cell, for $360$ probe--selector runs in total.

\Cref{tab:confounding_pertask} reports the fully confounded cell per task, separating false positives on confounded dimensions (FP$_{\mathrm{conf}}$, out of $12$) from those on exogenous distractors (FP$_{\mathrm{exo}}$). The failure of the observational selectors is specifically a confounding failure rather than general noise sensitivity: on \texttt{walker\_walk}, MI Select admits $11.8$ of the $12$ confounded dimensions, and multistep inverse dynamics admits $8.4$, whereas \IBD{} admits $0.2$. Variance Select is an instructive exception in the opposite direction: it admits almost no confounded dimensions but $15.4$ exogenous ones, because marginal variance is uninformative about the latent process, so its low F1 has a different origin from the other baselines'.

\begin{table}[h]
\centering
\small
\caption{Per-task results in the fully confounded cell ($\kappa = 0.5$, $\lambda = 1$; mean $\pm$ std over $5$ seeds). FP$_{\mathrm{conf}}$: confounded dimensions admitted (out of $12$). FP$_{\mathrm{exo}}$: exogenous distractors admitted. \textbf{Bold:} best F1 per task.}
\label{tab:confounding_pertask}
\setlength{\tabcolsep}{5pt}
\begin{tabular}{ll ccc}
\toprule
Task & Method & F1 & FP$_{\mathrm{conf}}$ & FP$_{\mathrm{exo}}$ \\
\midrule
\multirow{6}{*}{\texttt{walker\_walk}}
 & \IBD{} (ours) & \textbf{0.917}$\pm$0.035 & 0.2 & 2.0 \\
 & Inverse Dyn.  & 0.650$\pm$0.020 & 8.4 & 0.0 \\
 & MI Select     & 0.442$\pm$0.020 & 11.8 & 1.6 \\
 & Grad.\ Attr.  & 0.308$\pm$0.020 & 8.4 & 8.2 \\
 & Cond.\ MI     & 0.350$\pm$0.020 & 2.6 & 13.0 \\
 & Var Select    & 0.358$\pm$0.020 & 0.0 & 15.4 \\
\midrule
\multirow{6}{*}{\texttt{cheetah\_run}}
 & \IBD{} (ours) & \textbf{0.961}$\pm$0.022 & 0.0 & 1.4 \\
 & Inverse Dyn.  & 0.882$\pm$0.000 & 1.8 & 0.2 \\
 & MI Select     & 0.435$\pm$0.096 & 9.6 & 0.0 \\
 & Grad.\ Attr.  & 0.529$\pm$0.000 & 8.0 & 0.0 \\
 & Cond.\ MI     & 0.682$\pm$0.080 & 2.6 & 2.8 \\
 & Var Select    & 0.412$\pm$0.000 & 4.2 & 5.8 \\
\midrule
\multirow{6}{*}{\texttt{reacher\_hard}}
 & \IBD{} (ours) & \textbf{0.969}$\pm$0.038 & 0.0 & 0.4 \\
 & Inverse Dyn.  & 0.667$\pm$0.000 & 0.0 & 2.0 \\
 & MI Select     & 0.333$\pm$0.000 & 4.0 & 0.0 \\
 & Grad.\ Attr.  & 0.533$\pm$0.067 & 2.8 & 0.0 \\
 & Cond.\ MI     & 0.333$\pm$0.000 & 0.6 & 3.4 \\
 & Var Select    & 0.333$\pm$0.000 & 0.0 & 4.0 \\
\bottomrule
\end{tabular}
\end{table}

\section{Summary-Statistic Ablation}
\label{app:summary_stat}

The summary $\Delta^h_i$ of \Cref{eq:summary} measures step-to-step displacement, so it is by construction insensitive to an action effect that shifts the mean level of a dimension without changing its increments. This appendix quantifies both how much that blind spot costs on the standard benchmark and how it can be removed.

We compare three statistics, all scored on the same collected trajectories so that the comparison is exactly paired. Delta is the published $\Delta^h_i$; level is the per-trajectory mean of the raw values $o^{(i)}$; both takes the union of the two families of tests under a joint Benjamini--Hochberg correction, so the reported gains already pay the multiplicity cost of the extra family.

On four standard-benchmark settings (\texttt{cheetah\_run} medium, \texttt{walker\_walk} medium and hard, \texttt{hopper\_hop} medium), the displacement statistic dominates the level statistic everywhere, with recall higher by $0.35$--$0.49$, confirming that it is the appropriate default when every action-reachable dimension carries an interventional signal in its increments. Adding the level family raises recall where some dimensions respond weakly (\texttt{hopper\_hop} $0.773 \to 0.880$, \texttt{walker\_walk} medium $0.875 \to 0.950$) and costs a few points of precision where delta already achieves recall $1.0$ (\texttt{cheetah\_run}), since the second family then only adds opportunities for false positives.

The picture changes when the blind spot is actually exercised. We add $6$ dimensions that are action-reachable only through their mean level, leaving their increment distribution unchanged under intervention. \Cref{tab:levelshift} reports the detection rate on those constructed dimensions alongside overall boundary quality. The displacement statistic misses them almost entirely, recovering none of the six on \texttt{reacher\_hard}; the level statistic recovers all of them but loses the dimensions whose signal is carried by increments; the union recovers both families and attains the best overall F1 in all three tasks.

\begin{table}[h]
\centering
\small
\caption{Detection of level-shift dimensions (mean $\pm$ std over $5$ seeds, medium distractors, $6$ constructed dimensions per task). Level recall is the detection rate on the constructed dimensions; precision, recall, and F1 are over the full boundary.}
\label{tab:levelshift}
\begin{tabular}{ll cccc}
\toprule
Setting & Statistic & Level recall & Precision & Recall & F1 \\
\midrule
\multirow{3}{*}{\texttt{cheetah\_run}}
 & delta & 0.067$\pm$0.082 & 0.956$\pm$0.022 & 0.757$\pm$0.021 & 0.845$\pm$0.020 \\
 & level & 1.000$\pm$0.000 & 0.958$\pm$0.055 & 0.548$\pm$0.044 & 0.695$\pm$0.039 \\
 & both  & 1.000$\pm$0.000 & 0.936$\pm$0.030 & 1.000$\pm$0.000 & \textbf{0.967}$\pm$0.016 \\
\midrule
\multirow{3}{*}{\texttt{reacher\_hard}}
 & delta & 0.000$\pm$0.000 & 0.893$\pm$0.096 & 0.500$\pm$0.000 & 0.639$\pm$0.025 \\
 & level & 1.000$\pm$0.000 & 0.931$\pm$0.090 & 0.583$\pm$0.000 & 0.715$\pm$0.028 \\
 & both  & 1.000$\pm$0.000 & 0.905$\pm$0.090 & 1.000$\pm$0.000 & \textbf{0.948}$\pm$0.050 \\
\midrule
\multirow{3}{*}{\texttt{walker\_walk}}
 & delta & 0.033$\pm$0.067 & 0.911$\pm$0.028 & 0.740$\pm$0.033 & 0.816$\pm$0.022 \\
 & level & 1.000$\pm$0.000 & 0.962$\pm$0.047 & 0.327$\pm$0.013 & 0.488$\pm$0.019 \\
 & both  & 1.000$\pm$0.000 & 0.919$\pm$0.054 & 0.947$\pm$0.027 & \textbf{0.932}$\pm$0.034 \\
\bottomrule
\end{tabular}
\end{table}

Since the union costs at most a few points of precision when the blind spot is absent and is necessary when it is present, we recommend it as the default in deployments where the mechanism generating each dimension is unknown, and retain the displacement statistic alone for the results reported in the main text.

\section{Multiple-Testing and Test-Statistic Variants}
\label{app:multiplicity}

Proposition~\ref{prop:type1} invokes the standard Benjamini--Hochberg guarantee, which requires independence or PRDS among the dimension-level $p$-values. Mimicking distractors are calibrated against true proprioceptive channels and are therefore statistically coupled to them, so this condition cannot be taken for granted. We evaluate two axes on the confounded benchmark of \Cref{sec:confounding}: the two-sample test (Welch $t$-test versus a two-sample Kolmogorov--Smirnov test, which is sensitive to distributional changes beyond the mean) and the multiplicity correction (BH versus Benjamini--Yekutieli~\citep{benjamini2001dependency}, which is valid under arbitrary dependence at the cost of a $\sum_{i=1}^d 1/i$ factor). All four variants are computed from the same probe trajectories, so the comparison is exactly paired.

\begin{table}[h]
\centering
\small
\caption{Test-statistic and multiplicity variants, pooled over the four confounding cells, $3$ tasks, and $5$ seeds ($n = 60$). FDR is the realized false discovery proportion at nominal level $\alpha = 0.05$.}
\label{tab:multiplicity}
\begin{tabular}{lccccc}
\toprule
Variant & Precision & Recall & F1 & Realized FDR & FP$_{\mathrm{conf}}$ \\
\midrule
Welch + BH (default) & 0.913 & \textbf{0.975} & 0.942 & 0.087 & 0.23 \\
Welch + BY           & \textbf{1.000} & 0.954 & \textbf{0.975} & \textbf{0.000} & \textbf{0.00} \\
KS + BH              & 0.946 & 0.969 & 0.956 & 0.054 & 0.07 \\
KS + BY              & 0.993 & 0.938 & 0.963 & 0.007 & \textbf{0.00} \\
\bottomrule
\end{tabular}
\end{table}

Two observations follow from \Cref{tab:multiplicity}. First, the realized false discovery proportion under BH is $0.087$ against a nominal $\alpha = 0.05$, so the dependence among dimension-level $p$-values is strong enough to matter, consistent with the concern that the PRDS condition may not hold. Second, the remedy is inexpensive: the Benjamini--Yekutieli correction, which is valid under arbitrary dependence, drives the realized FDR to zero and admits no confounded dimension at all, while its more conservative threshold costs only $0.021$ recall. Its F1 is accordingly higher than the default ($0.975$ versus $0.942$). The KS test behaves similarly to Welch on this benchmark, indicating that the results do not lean on the specific choice of two-sample test.

We retain Welch + BH as the default so that the main results correspond to a single fixed configuration, and note that BY is the appropriate choice when a conservative guarantee is required and a small loss of recall is acceptable.



\end{document}

%% file: table1_main.tex
\begin{table}[htbp]
\centering
\caption{Main results (episode return, mean $\pm$ std over 5 seeds). \textbf{Bold}: best non-oracle method (within 1 std).}
\label{tab:main}
\resizebox{\textwidth}{!}{
\small
\setlength{\tabcolsep}{10pt}
\begin{tabular}{llrrrrrr}
\toprule
Environment & Distr. & Full State & \textbf{IBD (ours)} & Oracle & MI Select & Var Select & Cond.\ MI \\
\midrule
walker\_walk & easy & 857$\pm$139 & 854$\pm$127 & 785$\pm$128 & 705$\pm$98 & 750$\pm$103 & \textbf{884$\pm$70} \\
walker\_walk & medium & \textbf{902$\pm$26} & 792$\pm$112 & 785$\pm$128 & 129$\pm$79 & 84$\pm$6 & 220$\pm$135 \\
walker\_walk & hard & 414$\pm$144 & \textbf{842$\pm$121} & 785$\pm$128 & 87$\pm$34 & 46$\pm$10 & 81$\pm$23 \\
\addlinespace[3pt]
cheetah\_run & easy & \textbf{508$\pm$39} & 429$\pm$80 & 472$\pm$35 & 161$\pm$42 & 85$\pm$19 & 270$\pm$112 \\
cheetah\_run & medium & 113$\pm$34 & \textbf{479$\pm$95} & 472$\pm$35 & 38$\pm$11 & 62$\pm$25 & 70$\pm$37 \\
cheetah\_run & hard & 63$\pm$12 & \textbf{436$\pm$70} & 472$\pm$35 & 33$\pm$6 & 39$\pm$3 & 62$\pm$41 \\
\addlinespace[3pt]
reacher\_hard & easy & 748$\pm$125 & \textbf{939$\pm$38} & 925$\pm$75 & 197$\pm$375 & 4$\pm$2 & 476$\pm$289 \\
reacher\_hard & medium & 12$\pm$7 & \textbf{929$\pm$76} & 925$\pm$75 & 12$\pm$5 & 7$\pm$5 & 7$\pm$6 \\
reacher\_hard & hard & 8$\pm$4 & \textbf{907$\pm$72} & 925$\pm$75 & 13$\pm$9 & 10$\pm$5 & 9$\pm$5 \\
\addlinespace[3pt]
cartpole\_swingup & medium & 725$\pm$48 & \textbf{834$\pm$26} & 821$\pm$35 & 78$\pm$25 & 83$\pm$23 & 161$\pm$60 \\
\addlinespace[3pt]
finger\_spin & medium & 365$\pm$13 & \textbf{587$\pm$178} & 775$\pm$181 & 1$\pm$1 & 1$\pm$1 & 4$\pm$6 \\
\addlinespace[3pt]
hopper\_hop & medium & 0$\pm$1 & \textbf{6$\pm$12} & 6$\pm$8 & 0$\pm$0 & 0$\pm$0 & 0$\pm$0 \\
\bottomrule
\end{tabular}}
\vspace{-5mm}
\end{table}

%% file: table3_boundary.tex
\begin{table}[H]
\centering
\caption{IBD boundary discovery accuracy (P/R/F1, mean over 5 seeds). High precision = few distractor dims leaked; high recall = few true dims missed.}
\label{tab:boundary}
\begin{tabular}{ll ccc}
\toprule
Environment & Distr. & Precision & Recall & F1 \\
\midrule
walker\_\,walk & easy & 0.98 & 0.92 & 0.95 \\
walker\_\,walk & medium & 0.95 & 0.86 & 0.90 \\
walker\_\,walk & hard & 0.97 & 0.85 & 0.91 \\
cheetah\_\,run & easy & 0.98 & 1.00 & 0.99 \\
cheetah\_\,run & medium & 0.94 & 1.00 & 0.97 \\
cheetah\_\,run & hard & 0.96 & 1.00 & 0.98 \\
reacher\_\,hard & easy & 0.94 & 1.00 & 0.97 \\
reacher\_\,hard & medium & 0.97 & 1.00 & 0.98 \\
reacher\_\,hard & hard & 0.86 & 1.00 & 0.93 \\
cartpole\_\,swingup & medium & 0.93 & 1.00 & 0.96 \\
finger\_\,spin & medium & 0.96 & 1.00 & 0.98 \\
hopper\_\,hop & medium & 0.96 & 0.75 & 0.84 \\
\bottomrule
\end{tabular}
\end{table}